\documentclass[11pt]{article}

\usepackage[final]{acl}

\usepackage{times}
\usepackage{latexsym}

\usepackage[T1]{fontenc}

\usepackage[utf8]{inputenc}

\usepackage{microtype}

\usepackage{inconsolata}

\usepackage{graphicx}

\usepackage{times}
\usepackage{helvet}
\usepackage{courier}
\usepackage{amsmath, amssymb, amsthm, bm}
\newtheorem{proposition}{Proposition}
\newtheorem{lemma}{Lemma}
\newtheorem{theorem}{Theorem}
\newtheorem{corollary}{Corollary}
\usepackage{booktabs}
\usepackage{multirow}
\usepackage{graphicx}
\usepackage{hyperref}
\usepackage{url}
\usepackage{enumitem}
\usepackage{algorithm}
\usepackage{algpseudocode}
\usepackage{mathtools}
\usepackage{xcolor}
\usepackage{etoolbox}

\AtBeginDocument{%
  \setlength{\abovedisplayskip}{3pt plus 1pt minus 1pt}%
  \setlength{\belowdisplayskip}{3pt plus 1pt minus 1pt}%
  \setlength{\abovedisplayshortskip}{2pt plus 1pt minus 1pt}%
  \setlength{\belowdisplayshortskip}{2pt plus 1pt minus 1pt}%
}

\title{Muon-LoGo: Local Orthogonalization with On-Demand Global Corrections}

\author{Jiarun Shen \and Xiao Li \\
  The Chinese University of Hong Kong, Shenzhen \\
  \href{mailto:jiarunshen@link.cuhk.edu.cn}{\texttt{jiarunshen@link.cuhk.edu.cn}}\quad
  \href{mailto:lixiao@cuhk.edu.cn}{\texttt{lixiao@cuhk.edu.cn}}}

\hypersetup{
  pdftitle={Muon-LoGo: Local Orthogonalization with On-Demand Global Corrections},
  pdfauthor={Jiarun Shen, Xiao Li}
}

\begin{document}
\maketitle
\begin{abstract}
Muon orthogonalizes matrix updates, but full distributed orthogonalization
requires full-matrix communication when matrices are sharded across devices.
Blockwise Muon avoids this communication by orthogonalizing each shard independently, but ignores cross-shard structure.
The missing cross-shard structure is especially relevant to sparse mixture-of-experts (MoE) models, where MoE token routing changes update statistics across experts and training steps.
We present \textbf{Muon-LoGo} (Muon with on-demand LOcal-to-Global Orthogonalization), a method that chooses blockwise or full orthogonalization for each matrix.
Muon-LoGo monitors the coefficient of variation of spectral norms across tensor-parallel shards.
Muon-LoGo normalizes each observation using recent statistics for the same matrix and performs full orthogonalization only when the normalized score exceeds a threshold.
We establish norm control, endpoint recovery, and per-matrix convergence for adaptive geometry switching.
Under an explicit score-calibration assumption, we derive a threshold-dependent reference-error bound.
We also give exact communication accounting for independent matrix-level decisions.
Experiments on 0.4B-parameter dense and MoE language models show that Muon-LoGo improves over Blockwise Muon.
On the MoE model, Muon-LoGo achieves a final validation loss within 0.5\% of Full Muon while reducing full-orthogonalization communication by 65.0\%.~\footnote{Our code is available at \url{https://github.com/shenjiarun-2026/Megatron-LM/tree/adaptivemuon}.}
\end{abstract}

\begin{figure*}[t]
\centering
\includegraphics[width=0.92\textwidth]{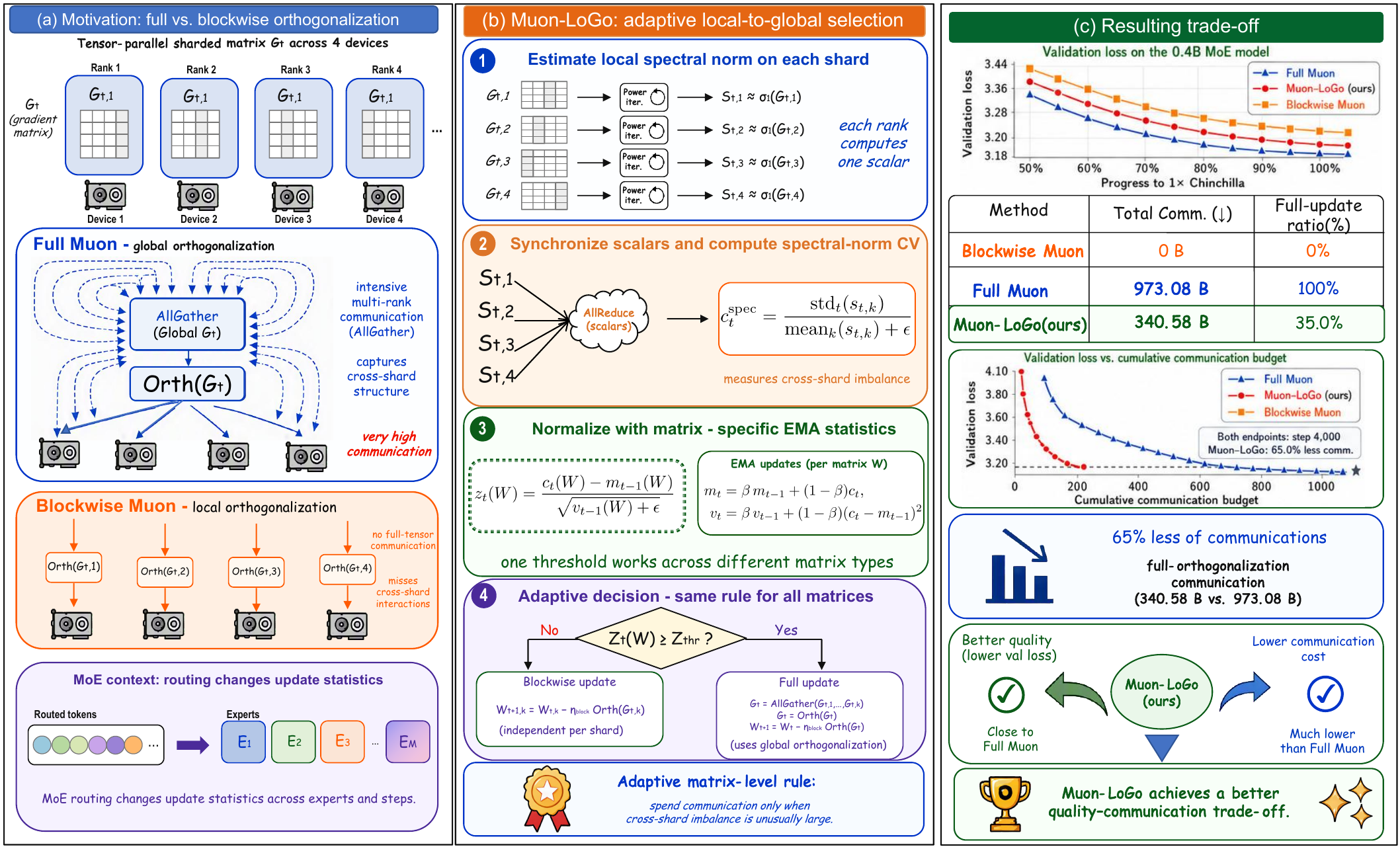}
\caption{Overview of Muon-LoGo. Full Muon captures cross-shard structure
at high communication cost, while Blockwise Muon uses local
orthogonalization. Muon-LoGo selects the orthogonalization scope for each
matrix using normalized cross-shard spectral-norm CV.}
\label{fig:method-overview}
\end{figure*}

\section{Introduction}

Orthogonalized optimizers offer a practical alternative to coordinate-wise methods for language model pretraining.
AdamW maintains coordinate-wise moments \cite{kingma2015adam,loshchilov2019adamw}, whereas Muon orthogonalizes momentum matrices and performs well in language model training \cite{jordan2024muon}.
Distributed Muon becomes costly when tensor parallelism partitions each matrix across devices.
Full Muon preserves cross-shard structure but communicates every eligible matrix at every step.
Blockwise Muon removes full-tensor matrix communication by orthogonalizing each local shard independently, but discards cross-shard interactions.
Kimi K3 applies Muon independently to attention heads
\cite{kimiteam2026kimik3}.
The per-head design avoids full-matrix orthogonalization across heads and
reduces the corresponding optimizer communication.

The choice between blockwise and full orthogonalization is important for sparse mixture-of-experts (MoE) models \cite{shazeer2017moe,lepikhin2021gshard,fedus2022switch}.
Distributed MoE training already requires communication for token
dispatch and combination. Dense feed-forward layers do not require token-to-expert communication.
Tensor parallelism also shards each expert matrix, so full
orthogonalization adds optimizer communication for every expert matrix.
MoE token routing changes update statistics across experts and training steps.
Fixed schedules cannot respond to such matrix-level variation.
An effective method must identify matrices that benefit from full orthogonalization without communicating every matrix.

We propose \textbf{Muon-LoGo}, a matrix-level method that selects blockwise or full orthogonalization from current shard statistics.
For each matrix, Muon-LoGo computes the coefficient of variation (CV) of spectral norms across tensor-parallel shards.
It normalizes the current CV with recent values for the same matrix and applies a full update when the normalized score exceeds a threshold.
The spectral norm captures the largest singular value of a shard, and its cross-shard CV increases when these values become more unequal.
Matrix-specific normalization permits one threshold across different parameter matrices.
Figure~\ref{fig:method-overview} summarizes the motivation, update
selection, and resulting quality and communication trade-off.

Our contributions are as follows.
\begin{itemize}[leftmargin=1.5em]
    \item We introduce a matrix-level method that selects between blockwise and full Muon updates using synchronized scalar spectral-norm estimates and matrix-specific normalization. Full-tensor communication is used only for selected matrices.
    \item We establish endpoint recovery, norm control, and per-matrix convergence for adaptive geometry switching. Under an explicit score-calibration assumption, we derive a threshold-dependent reference-error bound. An exact communication identity covers independent matrix-level decisions.
    \item Matched-budget experiments show improvements over Blockwise Muon on dense and MoE models. Controlled ablations compare Muon-LoGo with a model-level periodic baseline and compare spectral-norm CV with three alternative signals. On the MoE model, Muon-LoGo reduces full-orthogonalization communication by 65.0\% relative to Full Muon.
\end{itemize}

\section{Related Work}

\noindent\textbf{Orthogonalized optimizers.}
Muon introduced orthogonalized momentum for matrix-valued neural network parameters \cite{jordan2024muon}.
Recent work studies adaptive normalization and lower-cost distributed variants \cite{khaled2025muonbp,ahn2025dion,li2025normuon}.
Muon has also been used in large-scale training.
Kimi K2 uses MuonClip, Kimi K3 uses a per-head Muon variant, and DeepSeek-V4 applies Muon to most model parameters \cite{kimiteam2025kimik2,kimiteam2026kimik3,deepseekai2026deepseekv4}.
Muon-LoGo instead decides which matrices receive full distributed orthogonalization at each step.

\noindent\textbf{Distributed training systems.}
Tensor parallelism  partitions matrices across ranks.
Full Muon must reconstruct cross-shard structure and redistribute each orthogonalized update, adding optimizer communication and synchronization \cite{khaled2025muonbp,boreiko2025understanding}.
MuonBP uses periodic full updates \cite{khaled2025muonbp}, and Dion redesigns orthogonalized updates for distributed training \cite{ahn2025dion}.
Kimi K3 partitions attention-projection momentum by head and orthogonalizes each head block independently \cite{kimiteam2026kimik3}.
Fixed decompositions and schedules cannot adapt to current matrix statistics.
Muon-LoGo selects the update type online for each shard matrix.

\noindent\textbf{Sparse MoE training.}
MoE models increase capacity through sparse token-to-expert routing \cite{shazeer2017moe,lepikhin2021gshard,fedus2022switch}.
Token routing communicates activations, while tensor parallelism shards
each expert matrix and adds separate optimizer communication for full
orthogonalization. Routing also changes update statistics across experts
and steps. Muon-LoGo selects the orthogonalization scope from current
shard statistics.

\section{Background and Motivation}
\label{sec:background}

Muon orthogonalizes each matrix-valued gradient or momentum matrix $G_t$ before applying the update.
Newton--Schulz iterations approximate the orthogonalization in practice.
Tensor parallelism splits $G_t$ across ranks, so a full Muon update must
reconstruct the matrix, orthogonalize it, and return each shard.
In MoE training, full orthogonalization across sharded expert matrices
adds optimizer communication to the separate cost of token routing.

Blockwise Muon avoids full-tensor communication by orthogonalizing each local shard independently.
Let $H_{t,k}$ denote a momentum shard in Megatron's stored
$[\mathrm{out},\mathrm{in}]$ orientation.
Set $G_{t,k}=H_{t,k}$ for RowParallelLinear, which shards the input
dimension. Set $G_{t,k}=H_{t,k}^{\top}$ for ColumnParallelLinear, which
shards the output dimension.
The resulting canonical partition is
\begin{equation}
G_t = [G_{t,1}, G_{t,2}, \dots, G_{t,K}],
\end{equation}
The ColumnParallelLinear update is transposed back before it is applied.
Transposition preserves the spectral norm and the corresponding polar
update.
Full Muon computes $\text{Orth}(G_t)$, whereas blockwise Muon computes $\text{Orth}(G_{t,k})$ independently on each shard.
Full orthogonalization uses one normalization derived from all shards, whereas blockwise orthogonalization normalizes each shard separately.

\paragraph{Local orthogonalization.}
Kimi K3 orthogonalizes attention-projection momentum independently for
each head to reduce communication \cite{kimiteam2026kimik3}.
Kimi K3 uses a fixed head decomposition, whereas Muon-LoGo selects local
or full orthogonalization for each matrix from current cross-shard
statistics.

\paragraph{Selective full updates.}
Muon-LoGo synchronizes scalar shard statistics and communicates full
tensors only for selected matrices. Unlike periodic schedules, the
selection directs communication using each matrix's observed cross-shard
imbalance.

\section{Method}
\label{sec:method}

\subsection{Spectral-norm imbalance signal}

Muon-LoGo (Muon with on-demand LOcal-to-Global Orthogonalization) compares the largest singular values of tensor-parallel shards.
Each rank computes one scalar, and the ranks synchronize only scalar moments.

Starting from a unit vector $v_{t,k}^{(0)}$, we use $L$ power iterations:
\begin{equation}
\begin{aligned}
u_{t,k}^{(\ell)}
&=
\frac{G_{t,k}v_{t,k}^{(\ell-1)}}
{\|G_{t,k}v_{t,k}^{(\ell-1)}\|_2+\epsilon},
\\
v_{t,k}^{(\ell)}
&=
\frac{G_{t,k}^{\top}u_{t,k}^{(\ell)}}
{\|G_{t,k}^{\top}u_{t,k}^{(\ell)}\|_2+\epsilon}.
\end{aligned}
\end{equation}
The local spectral-norm estimate is
\begin{equation}
s_{t,k}
=
\|G_{t,k}v_{t,k}^{(L)}\|_2
\approx
\sigma_1(G_{t,k}).
\end{equation}
We use $L=2$ in all experiments.
Each iteration requires only the two local matrix-vector products
$G_{t,k}v$ and $G_{t,k}^{\top}u$.
For ColumnParallelLinear, these products use $H_{t,k}^{\top}$ and
$H_{t,k}$ in the opposite order, with vector dimensions set by the local
shape.
The computation is inexpensive compared with the matrix-matrix products
in Newton--Schulz orthogonalization and requires no matrix communication.
We then define the spectral-norm CV as
\begin{equation}
 c_t^{\mathrm{spec}}
 =
 \frac{\operatorname{std}_k(s_{t,k})}
 {\operatorname{mean}_k(s_{t,k}) + \epsilon}.
\end{equation}
Each rank synchronizes only the scalar estimate.
The CV is zero when all shard spectral norms are equal and increases as
their largest singular values become more unequal.

\subsection{Matrix-specific normalization}

A single raw CV threshold is unsuitable because attention, MLP, MoE router, and expert matrices have different baseline shard imbalance.
Muon-LoGo therefore normalizes each matrix's CV using its recent values.

For a monitoring signal $c_t(W)$, we maintain an exponentially weighted mean and a variance proxy.
The normalized score uses the statistics available before observing $c_t(W)$.
\begin{equation}
 z_t(W)=
 \frac{c_t(W)-m_{t-1}(W)}
 {\sqrt{v_{t-1}(W)+\epsilon}}.
\end{equation}
After selecting the update type, the statistics are updated as
\begin{equation}
\begin{split}
m_t(W)
&=\beta m_{t-1}(W)+(1-\beta)c_t(W),\\
v_t(W)
&=\beta v_{t-1}(W)\\[-2pt]
&\quad +(1-\beta)\bigl(c_t(W)-m_{t-1}(W)\bigr)^2.
\end{split}
\end{equation}
The first two observations initialize the running statistics and set $z_t(W)=0$.
The resulting score measures the change from the matrix's recent shard imbalance.

\subsection{Adaptive update selection}

Algorithm~\ref{alg:muon-logo} summarizes the Muon-LoGo update rule.
Muon-LoGo selects the update type for each matrix according to $z_t(W)$.

\paragraph{Blockwise update.}
When $z_t < z_{\mathrm{thr}}$, the optimizer applies local blockwise orthogonalization.
\begin{equation}
 W_{t+1,k}
 =
 W_{t,k}-\eta_{\mathrm{block}}\,\operatorname{Orth}(G_{t,k}).
\end{equation}

\paragraph{Full update.}
When $z_t \ge z_{\mathrm{thr}}$, the optimizer gathers the full matrix, applies full orthogonalization, and scatters the update back.
\begin{equation}
\begin{aligned}
G_t&=\text{AllGather}(G_{t,1},\dots,G_{t,K}),\\[-2pt]
\widetilde{G}_t&=\operatorname{Orth}(G_t),
\end{aligned}
\end{equation}
\begin{equation}
 W_{t+1,k}=W_{t,k}-\eta_{\mathrm{full}}\,\widetilde{G}_{t,k}.
\end{equation}
The rule permits separate learning rates $\eta_{\mathrm{block}}$ and $\eta_{\mathrm{full}}$.
Our experiments use the same tuned base learning rate for both updates.

\begin{algorithm}[t]
\caption{Muon-LoGo}
\label{alg:muon-logo}
\begin{algorithmic}[1]
\Require TP degree $K$, local shard $G_{t,k}$, learning rates $\eta_{\mathrm{block}},\eta_{\mathrm{full}}$, EMA decay $\beta$, warmup $T_{\mathrm{warm}}$, threshold $z_{\mathrm{thr}}$, constant $\epsilon$
\For{each training step $t$ and each TP-sharded matrix $W$}
    \If{$t<T_{\mathrm{warm}}$}
        \State Apply local blockwise orthogonalization with $\eta_{\mathrm{block}}$
        \State \textbf{continue}
    \EndIf
    \For{each rank $k \in \{1,\dots,K\}$}
        \State Estimate $s_{t,k}\approx\sigma_1(G_{t,k})$ with $L$ power iterations
    \EndFor
    \State Synchronize scalars and compute
    \[
    c_t(W)\gets \frac{\operatorname{std}_k(s_{t,k})}{\operatorname{mean}_k(s_{t,k})+\epsilon}
    \]
    \State Compute $z_t(W)\gets (c_t(W)-m_{t-1}(W))/\sqrt{v_{t-1}(W)+\epsilon}$
    \State Update $m_t(W)$ and $v_t(W)$ using $c_t(W)$
    \If{$z_t < z_{\mathrm{thr}}$}
        \State Apply local blockwise orthogonalization with $\eta_{\mathrm{block}}$
    \Else
        \State Gather the full matrix across TP ranks
        \State Apply full orthogonalization
        \State Scatter the update and apply it with $\eta_{\mathrm{full}}$
    \EndIf
\EndFor
\end{algorithmic}
\end{algorithm}

\subsection{Communication mechanisms}
\label{sec:communication-mechanisms}

The three Muon variants differ in how they convert tensor-parallel momentum shards into an orthogonalized update.
All variants share the forward and backward collectives and standard gradient synchronization.
\textbf{Blockwise Muon} orthogonalizes each local shard independently.
It introduces no cross-rank matrix transfer but uses a separate Gram
matrix $G_{t,k}G_{t,k}^{\top}$ on each shard rather than the aggregate
$\sum_k G_{t,k}G_{t,k}^{\top}$ used by full orthogonalization.
For the full matrix $H_t$ in stored orientation, the aggregate is
$H_tH_t^{\top}$ for RowParallelLinear and $H_t^{\top}H_t$ for
ColumnParallelLinear.

\textbf{Full Muon} reconstructs the global matrix for every eligible parameter at every step.
Each update therefore requires full-tensor interaction within the tensor-parallel group.

\textbf{Muon-LoGo} selects the update type for each matrix.
It first synchronizes scalar shard statistics.
A blockwise update follows Blockwise Muon, while a full update follows
Full Muon for that matrix.
The same optimizer step may contain both update types.
Communication depends on the total size of matrices assigned full updates.
Proposition~\ref{prop:communication-accounting} gives the exact size-weighted cost.
Appendix~\ref{app:communication-execution} describes the post-backward execution.

\section{Theoretical Analysis}

The geometric and convergence results analyze one parameter matrix with
one decision $a_t$ per step. The result covers adaptive geometry
switching, not a product space with independent $a_{t,i}$ across
matrices. The communication result covers the matrix-level decisions
used by Muon-LoGo.
The appendix provides complete statements and proofs.

\subsection{Special cases and geometric interpretation}

Muon-LoGo uses blockwise orthogonalization as a low-cost approximation to Full Muon.
The threshold defines a family of update rules between Blockwise Muon and Full Muon.

\begin{proposition}[Special-case recovery]
\label{prop:special-cases}
Consider Muon-LoGo after the local-update warmup.
If $z_{\mathrm{thr}}\rightarrow +\infty$, Muon-LoGo selects only blockwise updates and reduces to Blockwise Muon.
If $z_{\mathrm{thr}}\rightarrow -\infty$, every post-warmup update is full.
With zero warmup, the second limit recovers Full Muon for the entire trajectory.
\end{proposition}

The two limits recover the standard Muon variants.

\subsection{Geometric control}

Lemma~\ref{lem:norm-control} bounds full and blockwise polar directions
under the operator and block-spectral geometries.
Lemma~\ref{lem:monitor-control} assumes an external nondecreasing
function $\phi$ that calibrates the score to the directional difference.
A blockwise decision then adds at most $\phi(z_{\mathrm{thr}})$ reference
error beyond finite-step orthogonalization error. The lemma does not show
that spectral-norm CV controls the directional difference.

\newcommand{\DetailedConvergenceResults}{%
\subsection{Per-matrix convergence}

We state a per-matrix mixed-geometry convergence result.
The analysis treats the method as an inexact linear-minimization procedure with geometry selected by $a_t$.
The bound separates descent from the selected exact oracle, error between the momentum matrix and the true gradient, and finite-step orthogonalization error.

\paragraph{Notation.}
Let the canonical tensor-parallel partition of one matrix $X$ be
$X=[X_1,\ldots,X_K]$.
For a stored row partition, $X$ denotes the transpose defined in
Section~\ref{sec:background}.
We define the full operator norm and the block-spectral norm by
\begin{equation}
\begin{aligned}
\|X\|_{\mathrm{F}}
&:=
\|X\|_{\mathrm{op}},
\\
\|X\|_{\mathrm{B}}
&:=
\max_{1\le b\le K}\|X_b\|_{\mathrm{op}}.
\end{aligned}
\end{equation}
Their dual norms are
\begin{equation}
\begin{aligned}
\|Y\|_{\mathrm{F},*}
&=
\|Y\|_{*},
\\
\|Y\|_{\mathrm{B},*}
&=
\sum_{b=1}^{K}\|Y_b\|_{*},
\end{aligned}
\end{equation}
where $\|\cdot\|_{*}$ denotes the nuclear norm. In particular,
\begin{equation}
\|Y\|_{\mathrm{F},*}
\le
\|Y\|_{\mathrm{B},*}.
\end{equation}
The subscript $\mathrm{F}$ denotes the full-matrix geometry rather than the Frobenius norm.
All inner products are Frobenius inner products.

Let
\begin{equation}
a_t
=
\mathbf{1}\!\left[z_t\ge z_{\mathrm{thr}}\right]
\in\{0,1\}
\end{equation}
denote the update-selection decision, where $a_t=1$ selects a full Muon
update and $a_t=0$ selects a blockwise Muon update. Define
\begin{equation}
\begin{aligned}
j_t
&=
\begin{cases}
\mathrm{F}, & a_t=1,\\
\mathrm{B}, & a_t=0,
\end{cases}
\\
\eta_t
&=
a_t\eta_{\mathrm{full}}
+
(1-a_t)\eta_{\mathrm{block}},
\end{aligned}
\end{equation}
and
\begin{equation}
L_t
=
L_{j_t}.
\end{equation}

Let $M_t$ denote the stochastic momentum estimator used by Muon,
for example
\begin{equation}
M_t
=
\mu M_{t-1}
+
(1-\mu)G_t.
\end{equation}
For geometry $j\in\{\mathrm{F},\mathrm{B}\}$, define the exact
linear-minimization-oracle direction
\begin{equation}
U_t^{j}
\in
\operatorname*{arg\,max}_{\|U\|_{j}\le 1}
\langle M_t,U\rangle.
\end{equation}
Thus,
\begin{equation}
\begin{aligned}
U_t^{\mathrm{F}}
&=
\operatorname{Polar}(M_t),
\\
U_t^{\mathrm{B}}
&=
\big[
\operatorname{Polar}(M_{t,1}),
\dots,
\operatorname{Polar}(M_{t,K})
\big].
\end{aligned}
\end{equation}
The variational identity
\begin{equation}
\langle M_t,U_t^j\rangle
=
\|M_t\|_{j,*}
\end{equation}
holds for either geometry.

Let $\widehat U_t$ be the direction actually returned by
Muon-LoGo, including finite-step Newton--Schulz error. The
parameter update is
\begin{equation}
X_{t+1}
=
X_t-\eta_t\widehat U_t.
\end{equation}
Any deterministic shape-dependent scaling applied after
orthogonalization can be absorbed into $\eta_t$; the analysis therefore
uses unit-norm oracle directions without loss of generality.

\begin{theorem}[Per-matrix adaptive mixed-geometry convergence]
\label{thm:muon-logo-convergence}
Suppose that $f:\mathbb{R}^{m\times n}\rightarrow\mathbb{R}$ is
lower bounded by $f_\star$, and let
\begin{equation}
\Delta_0
=
f(X_0)-f_\star.
\end{equation}
Assume that $f$ is smooth under both geometries.
For every
$j\in\{\mathrm{F},\mathrm{B}\}$ and all $X,Y$,
\begin{equation}
\begin{aligned}
f(Y)
&\le
f(X)
+
\langle\nabla f(X),Y-X\rangle
\\
&\quad +
\frac{L_j}{2}\|Y-X\|_j^2,
\end{aligned}
\end{equation}
where
\begin{equation}
L_{\mathrm{F}}:=L_{\mathrm{op}},
\qquad
L_{\mathrm{B}}>0.
\end{equation}

Assume that there are nonnegative, possibly data-dependent sequences
$\{\nu_t\}_{t\ge0}$ and $\{\delta_t\}_{t\ge0}$ such that, almost
surely,
\begin{equation}
\|M_t-\nabla f(X_t)\|_{j_t,*}
\le
\nu_t,
\end{equation}
and
\begin{equation}
\|\widehat U_t-U_t^{j_t}\|_{j_t}
\le
\delta_t,
\qquad
0\le\delta_t<1.
\end{equation}

Then, for any adaptive update-selection sequence
$\{a_t\}_{t=0}^{T-1}$ and any positive stepsizes
$\{\eta_t\}_{t=0}^{T-1}$, the iterates of Muon-LoGo satisfy,
almost surely,
\begin{equation}
\begin{aligned}
&\min_{0\le t<T}
\|\nabla f(X_t)\|_{*}
\\
&\quad \le
\frac{
\begin{gathered}
\displaystyle
\Delta_0+2\sum_{t=0}^{T-1}\eta_t\nu_t\\[-2pt]
\displaystyle
+\frac{1}{2}\sum_{t=0}^{T-1}
L_t\eta_t^2(1+\delta_t)^2
\end{gathered}
}{
\displaystyle
\sum_{t=0}^{T-1}
\eta_t(1-\delta_t)
}.
\end{aligned}
\label{eq:mixed-min}
\end{equation}

More precisely, the following weighted stationarity bound holds.
\begin{equation}
\begin{aligned}
&
\sum_{t=0}^{T-1}
\eta_t(1-\delta_t)
\|\nabla f(X_t)\|_{*}
\\[-2pt]
&\quad\le
\Delta_0
+
2\sum_{t=0}^{T-1}\eta_t\nu_t
\\[-2pt]
&\qquad+
\frac{1}{2}
\sum_{t=0}^{T-1}
L_t\eta_t^2(1+\delta_t)^2.
\end{aligned}
\label{eq:mixed-cumulative}
\end{equation}

Moreover, suppose that the following conditions hold almost surely.
\begin{equation}
\begin{aligned}
&\sum_{t=0}^{\infty}\eta_t=\infty,
\qquad
\sum_{t=0}^{\infty}\eta_t^2<\infty,
\\
&\sup_t\delta_t<1,
\qquad
\frac{
\sum_{t=0}^{T-1}\eta_t\nu_t
}{
\sum_{t=0}^{T-1}\eta_t
}
\longrightarrow 0,
\end{aligned}
\end{equation}
Then
\begin{equation}
\liminf_{t\rightarrow\infty}
\|\nabla f(X_t)\|_{*}
=
0.
\end{equation}
\end{theorem}

\begin{corollary}[Exact-oracle full/block interpolation]
\label{cor:muon-logo-interpolation}
Suppose $\nu_t=\delta_t=0$ for all $t$. Let
\begin{equation}
\begin{aligned}
N_{\mathrm{full}}
&=
\sum_{t=0}^{T-1}a_t,
\qquad
N_{\mathrm{block}}
=
T-N_{\mathrm{full}},
\\
r_T
&=
\frac{N_{\mathrm{full}}}{T}.
\end{aligned}
\end{equation}
Choose the geometry-dependent stepsizes
\begin{equation}
\eta_{\mathrm{full}}
=
\frac{c}{L_{\mathrm{op}}},
\qquad
\eta_{\mathrm{block}}
=
\frac{c}{L_{\mathrm{B}}},
\end{equation}
where $c>0$. Then
\begin{equation}
\begin{aligned}
\min_{0\le t<T}
\|\nabla f(X_t)\|_{*}
&\le
\frac{\Delta_0}{
c\left(
N_{\mathrm{full}}/L_{\mathrm{op}}
+
N_{\mathrm{block}}/L_{\mathrm{B}}
\right)}
\\[-2pt]
&\quad+
\frac{c}{2}.
\end{aligned}
\end{equation}
The optimal value is attained at
\begin{equation}
c^\star
=
\sqrt{
\frac{
2\Delta_0
}{
N_{\mathrm{full}}/L_{\mathrm{op}}
+
N_{\mathrm{block}}/L_{\mathrm{B}}
}
},
\end{equation}
which yields
\begin{equation}
\begin{aligned}
&\min_{0\le t<T}
\|\nabla f(X_t)\|_{*}
\\
&\quad \le
\sqrt{
\frac{
2\Delta_0
}{
N_{\mathrm{full}}/L_{\mathrm{op}}
+
N_{\mathrm{block}}/L_{\mathrm{B}}
}
}.
\end{aligned}
\label{eq:interpolation-counts}
\end{equation}

Equivalently, define the realized effective smoothness
\begin{equation}
L_{\mathrm{eff}}(r_T)
=
\left(
\frac{r_T}{L_{\mathrm{op}}}
+
\frac{1-r_T}{L_{\mathrm{B}}}
\right)^{-1}.
\end{equation}
Then
\begin{equation}
\begin{aligned}
&\min_{0\le t<T}
\|\nabla f(X_t)\|_{*}
\\
&\quad \le
\sqrt{
\frac{
2\Delta_0L_{\mathrm{eff}}(r_T)
}{T}
}.
\end{aligned}
\label{eq:effective-smoothness}
\end{equation}

Consequently,
\begin{equation}
r_T=1
\quad\Longrightarrow\quad
L_{\mathrm{eff}}(r_T)=L_{\mathrm{op}},
\end{equation}
and the sequence recovers the Full Muon per-matrix bound
\begin{equation}
\min_{t<T}
\|\nabla f(X_t)\|_{*}
\le
\sqrt{
\frac{2\Delta_0L_{\mathrm{op}}}{T}
}.
\end{equation}
Similarly,
\begin{equation}
r_T=0
\quad\Longrightarrow\quad
L_{\mathrm{eff}}(r_T)=L_{\mathrm{B}},
\end{equation}
and the sequence recovers the Blockwise Muon per-matrix bound
\begin{equation}
\min_{t<T}
\|\nabla f(X_t)\|_{*}
\le
\sqrt{
\frac{2\Delta_0L_{\mathrm{B}}}{T}
}.
\end{equation}
If $L_{\mathrm{op}}\le L_{\mathrm{B}}$, then
$L_{\mathrm{eff}}(r_T)$ is non-increasing in $r_T$, so increasing
the realized full-update ratio weakly improves the upper bound.
\end{corollary}
}

\subsection{Convergence summary}

For one parameter matrix, Theorem~\ref{thm:muon-logo-convergence}
establishes convergence under operator- and block-geometry smoothness,
momentum error $\nu_t$, and orthogonalization error $\delta_t<1$.
Define
\begin{equation}
\mathcal{E}_T
=
2\sum_{t<T}\eta_t\nu_t
+
\frac12\sum_{t<T}L_t\eta_t^2(1+\delta_t)^2.
\end{equation}
The single-matrix sequence satisfies
\begin{equation}
\min_{0\le t<T}\|\nabla f(X_t)\|_*
\le
\frac{\Delta_0+\mathcal{E}_T}
{\sum_{t<T}\eta_t(1-\delta_t)}.
\end{equation}
Corollary~\ref{cor:muon-logo-interpolation} recovers the Blockwise Muon and Full Muon rates at the endpoint update rules.
Corollary~\ref{cor:muon-logo-reference-error} bounds error relative to the Full Muon direction.
For a fixed score sequence, Proposition~\ref{prop:threshold-tradeoff} shows that lowering the threshold increases full updates and decreases the calibrated blockwise-error bound.

\subsection{Communication accounting}

Throughout the accounting and experiments, communication is measured by
the number of logical full-tensor elements communicated by distributed
orthogonalization at each optimizer step. Summing the per-step counts
gives the cumulative communication.
The monitoring step adds a local spectral-norm estimate and scalar
collectives for each sharded matrix.
It does not require full-matrix communication.

\begin{proposition}[Exact size-weighted communication accounting]
\label{prop:communication-accounting}
Suppose update types are selected independently for $J$ tensor-parallel matrices over $T$
optimizer steps. Let $q_i$ be the logical full-tensor element count of
matrix $i$, let $a_{t,i}\in\{0,1\}$ indicate a full update, and
let $Q=\sum_{i=1}^{J}q_i$. The cumulative full-orthogonalization
logical communication is exactly
\begin{equation}
\mathrm{Comm}_{\mathrm{full}}(T)
=
\sum_{t=0}^{T-1}\sum_{i=1}^{J}a_{t,i}q_i
=
\rho_T^{\mathrm{w}}\,TQ,
\end{equation}
where
\begin{equation}
\rho_T^{\mathrm{w}}
=
\frac{\sum_{t=0}^{T-1}\sum_{i=1}^{J}a_{t,i}q_i}{TQ}
\in[0,1]
\end{equation}
is the size-weighted full-update ratio. Therefore the
average communication per optimizer step is
$\rho_T^{\mathrm{w}}Q$. Blockwise Muon has
$\rho_T^{\mathrm{w}}=0$, whereas Full Muon has
$\rho_T^{\mathrm{w}}=1$. The low-dimensional monitoring collectives are
additive and are not included in this full-orthogonalization measure.
\end{proposition}

The identity weights each full update by matrix size rather than treating all matrices or steps equally.
Appendix~\ref{app:communication-execution} gives the execution-level accounting.

\section{Experiments}

\subsection{Setup}

We evaluate 0.4B-parameter Qwen3-style dense and sparse MoE
decoder-only models.
The dense model has 24 Transformer layers, hidden size 768, feed-forward
hidden size 2304, 12 attention heads, and four query groups.
The MoE model activates 0.21B parameters and has 16 Transformer layers,
hidden size 512, eight attention heads, and four query groups.
Each Transformer layer uses an MoE feed-forward block with 48 experts,
an expert hidden size of 192, and top-8 routing.
Both models use SwiGLU, RMSNorm, QK normalization, RoPE with base
1,000,000, untied input and output embeddings, and bias-free linear
layers. The maximum position embedding length is 32,768.
MoE runs use sequence length 4096, global batch size 256, and 4,000 optimizer steps.
Dense runs use 7,629 steps.
For each model family, the training budget is $20\times$ the number of activated parameters.
All runs use BF16 and a common validation protocol within each model family.
Appendix~\ref{app:implementation-details} reports the system details.

\paragraph{Data and tokenization.}
All models are pretrained on the English C4 corpus from the
\href{https://huggingface.co/datasets/allenai/c4}{AllenAI release}
\cite{raffel2020exploring}.
We use the official training split for optimization and the validation
split for evaluation.
No test split is used.
C4 is distributed under the ODC-BY 1.0 license, while source web content retains its original terms.
We use the Qwen3 byte-level BPE tokenizer with 151,669 tokens and pad the
model vocabulary to 151,936 tokens \cite{qwen2025qwen3}.

\paragraph{Initialization.}
Models are trained from scratch with Megatron-LM's default initialization.
Embeddings and non-output weights are sampled from a zero-mean normal distribution with standard deviation 0.02.
Transformer output projections use standard deviation $0.02/\sqrt{2L}$ for $L$ layers.
All optimizers within a model family use the same initialization and data order.

In the MoE experiments, all Muon variants use momentum 0.98, Nesterov momentum, five Newton--Schulz steps, and unit-RMS scaling for matrix parameters.
Adam updates the remaining scalar parameters.
Muon uses a base learning rate of 0.015, minimum learning rate $1.5\times10^{-4}$, weight decay 0.05, cosine decay, and 400 warmup steps.
Blockwise and full updates share the learning rate.
Muon-LoGo uses EMA coefficient 0.98, a 200-step blockwise warmup, two power iterations, and threshold 0.3.
We select 0.3 because it gives the lowest validation loss in the sweep in
Table~\ref{tab:ablation-z-threshold}, reported in
Appendix~\ref{app:ablation-results}.
AdamW uses learning rate 0.002, minimum learning rate $2\times10^{-5}$, and weight decay 0.1 under the same budget and evaluation cadence.
\subsection{Hyperparameter tuning protocol}

For the dense model, we tune AdamW and Blockwise Muon in two stages using
configurations from similarly sized pretrained models
\cite{wen2025fantastic,vyas2025soap,jordan2024muon}.
The first stage searches a broad range, and the second refines the best setting.
All candidates share the data, token budget, batch size, schedule family, and evaluation cadence.
We select the configuration with the lowest endpoint validation loss.
The Muon variants share optimizer hyperparameters to isolate the update-selection strategy.
Appendix~\ref{app:sweep-results} reports the dense-model sweeps.

\subsection{Baselines and metrics}

We compare AdamW, Blockwise Muon, Full Muon, and Muon-LoGo.
Blockwise Muon orthogonalizes each tensor-parallel shard independently and incurs no full-orthogonalization communication.
Full Muon performs distributed orthogonalization for every eligible matrix at every step.
Muon-LoGo selectively replaces blockwise updates with full updates.

Endpoint validation loss and PPL come from the final validation evaluation.
Communication is the cumulative number of logical full-tensor elements
communicated by distributed orthogonalization, as defined in
Proposition~\ref{prop:communication-accounting}.
The count excludes standard training communication and scalar monitoring
collectives.

\subsection{Endpoint validation quality}

\begin{figure*}[t]
\centering
\includegraphics[width=0.78\textwidth]{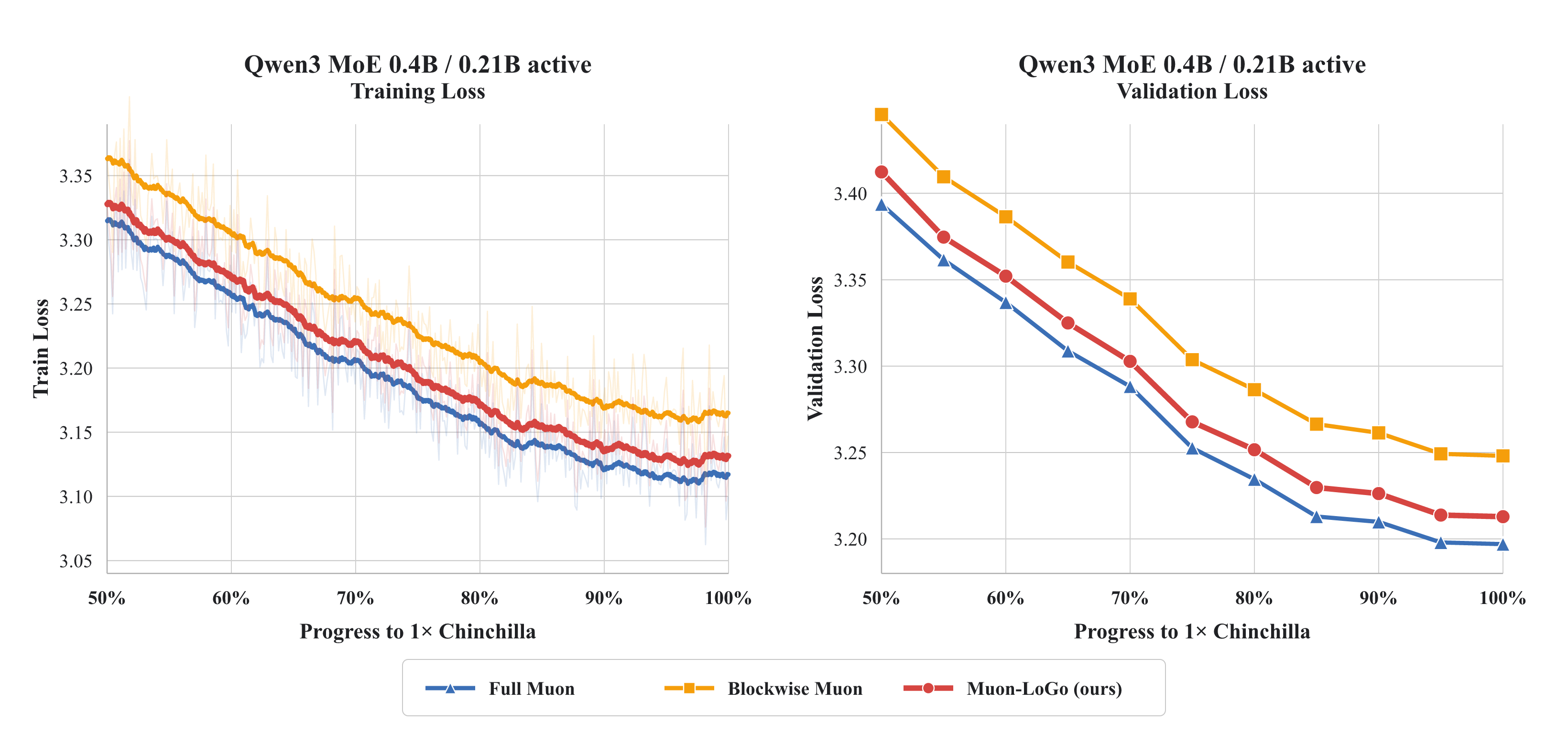}
\caption{Endpoint validation loss and PPL on the 0.4B MoE model after
4,000 optimizer steps. The plotted values are the results reported in
Table~\ref{tab:main-result}.}
\label{fig:validation-loss}
\end{figure*}

Figure~\ref{fig:validation-loss} compares endpoint validation quality
under the matched training budget. Full Muon has the lowest loss and
PPL among the three Muon variants, followed by Muon-LoGo and Blockwise
Muon.

\subsection{MoE quality and communication}

\begin{table*}[t]
\centering
\scriptsize
\setlength{\tabcolsep}{2pt}
\renewcommand{\arraystretch}{1.08}
\caption{Matched-budget results on the 0.4B MoE model after 4,000
optimizer steps (4.19B tokens). ``Avg. Comm.'' and ``Total Comm.'' count
logical matrix elements communicated by full orthogonalization. The
size-weighted full-update ratio is the quantity
$\rho_T^{\mathrm{w}}$ from
Proposition~\ref{prop:communication-accounting}.}
\label{tab:main-result}
\begin{tabular}{lcccccc}
\toprule
Optimizer
& Val Loss $\downarrow$
& Val PPL $\downarrow$
& First PPL $\le25.0$
& \shortstack{Avg. Comm. /\\Step (M) $\downarrow$}
& \shortstack{Total Comm.\\ (B) $\downarrow$}
& \shortstack{Size-Weighted\\Full-Update\\Ratio $\downarrow$} \\
\midrule
AdamW
& 3.3202
& 27.67
& not reached
& 0
& 0
& 0 \\

Blockwise Muon
& 3.2180
& 24.98
& 4,000 (final)
& 0
& 0
& 0 \\

Full Muon
& \textbf{3.1969}
& \textbf{24.46}
& \textbf{3,400}
& 243.27
& 973.08
& 100.0\% \\

Muon-LoGo
& 3.2128
& 24.85
& 3,800
& 85.14
& 340.58
& 35.0\% \\
\bottomrule
\end{tabular}
\end{table*}

\begin{figure*}[t]
\centering
\includegraphics[width=0.88\textwidth]{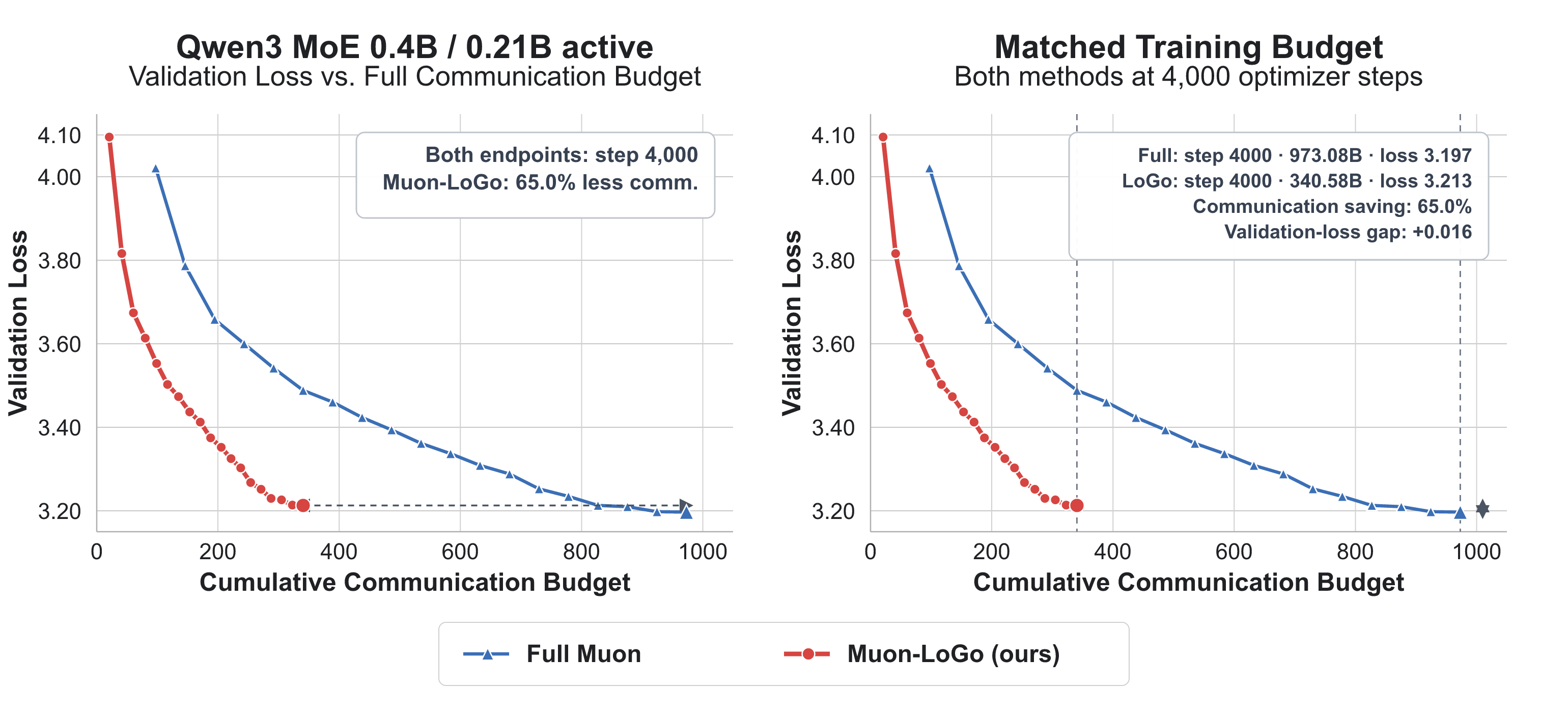}
\caption{Endpoint validation loss and aggregate full-orthogonalization
communication on the 0.4B MoE model. The left panel compares endpoint
quality with total communication. The right panel compares the total
logical matrix elements communicated by full orthogonalization. Muon-LoGo
uses 340.58B elements, 65.0\% less than Full Muon's 973.08B elements,
while its validation loss is 0.0159 higher. Blockwise Muon uses no
full-orthogonalization communication.}
\label{fig:communication-budget-validation-loss}
\end{figure*}

Table~\ref{tab:main-result} shows that Muon-LoGo improves Blockwise Muon
from 3.2180 to 3.2128 in validation loss and from 24.98 to 24.85 in PPL.
Muon-LoGo communicates 340.58B matrix elements, or 35.0\% of Full Muon's
973.08B elements.
Its final validation loss is 0.0159, or 0.5\%, above Full Muon.

Figure~\ref{fig:communication-budget-validation-loss} shows the endpoint
quality and aggregate communication reported in
Table~\ref{tab:main-result}. Muon-LoGo remains close to Full Muon in
validation loss while reducing full-orthogonalization communication.

\subsection{Results across dense and MoE models}

\begin{table*}[!t]
\centering
\caption{Endpoint quality and aggregate communication on the 0.4B dense
model after 7,629 optimizer steps (approximately 8B tokens).}
\label{tab:main-result-dense-model}
\scriptsize
\setlength{\tabcolsep}{1.5pt}
\renewcommand{\arraystretch}{1.00}
\begin{tabular}{lcccccc}
\toprule
Optimizer
& Val Loss $\downarrow$
& Val PPL $\downarrow$
& First PPL $\le25.0$
& \shortstack{Avg. Comm. /\\Step (M) $\downarrow$}
& \shortstack{Total Comm.\\ (B) $\downarrow$}
& \shortstack{Size-Weighted\\Full-Update\\Ratio $\downarrow$} \\
\midrule
AdamW & 3.220 & 25.03 & 6,000 & 0 & 0 & 0 \\
Blockwise Muon & 3.01 & 20.29 & 5,000 & 0 & 0 & 0 \\
Full Muon & \textbf{2.95} & \textbf{19.11} & \textbf{4,400} & 165.15 & 1255.15 & 100.0\% \\
Muon-LoGo & 2.99 & 19.89 & 4,900 & 48.94 & 371.98 & 29.64\% \\
\bottomrule
\end{tabular}
\end{table*}

On the dense model, Full Muon has the lowest validation loss, followed
by Muon-LoGo, Blockwise Muon, and AdamW.
Muon-LoGo improves Blockwise Muon from 3.01 to 2.99 in validation loss
and from 20.29 to 19.89 in PPL.
Full Muon reaches 2.95 loss and 19.11 PPL.
Together with the MoE results, the dense results show that matrix-level
selection improves the blockwise baseline on both models.

\section{Analysis}
\label{sec:communication-analysis}

\begin{table*}[!t]
\centering
\caption{Monitoring-signal ablation on the 0.4B MoE model after 4,000
optimizer steps. All runs use the same optimization setting and
normalized-score threshold. Communication follows
Proposition~\ref{prop:communication-accounting}.}
\label{tab:monitor-signal-main}
\scriptsize
\setlength{\tabcolsep}{3pt}
\renewcommand{\arraystretch}{1.10}
\begin{tabular*}{\textwidth}{@{\extracolsep{\fill}}lcccccc@{}}
\toprule
& \multicolumn{3}{c}{Validation} &
\multicolumn{3}{c}{Full-Orthogonalization Communication} \\
\cmidrule(lr){2-4}\cmidrule(l){5-7}
Signal & Loss $\downarrow$ & PPL $\downarrow$
& \shortstack{First PPL\\$\le25.0$}
& \shortstack{Avg. / Step\\(M) $\downarrow$}
& \shortstack{Total\\(B) $\downarrow$}
& \shortstack{Size-Weighted\\Full-Update Ratio $\downarrow$} \\
\midrule
Spectral-norm CV & \textbf{3.2128} & \textbf{24.85} & \textbf{3,800}
& \textbf{85.14} & \textbf{340.58} & \textbf{35.0\%} \\
Energy CV & 3.2533 & 25.88 & not reached & 91.54 & 366.18 & 37.6\% \\
Stable-rank CV & 3.2669 & 26.23 & not reached & 92.14 & 368.58 & 37.8\% \\
Directional Gram CV & 3.2995 & 27.10 & not reached & 92.94 & 371.78 & 38.2\% \\
\bottomrule
\end{tabular*}
\end{table*}

Muon-LoGo lies between Blockwise Muon and Full Muon in endpoint loss on
both models. It uses 35.0\% and 29.64\% of Full Muon's communication on
the MoE and dense models.

\subsection{Full-Update Ratios across Model Types}

Tables~\ref{tab:main-result} and~\ref{tab:main-result-dense-model} report
size-weighted full-update ratios of 35.0\% for the MoE model and 29.64\% for
the dense model. Thus, the same rule assigns more full updates to the
MoE model, while blockwise updates remain the majority for both models.

\subsection{How Much Full Orthogonalization Is Needed?}
\label{sec:ablations}

Across 128,000 matrix-level decisions, the selected setting uses 46,058
full updates and reaches validation loss within 0.5\% of Full Muon.
Table~\ref{tab:ablation-z-threshold} varies only the threshold and shows
that more full updates do not consistently improve validation quality.
The unweighted full-update ratio decreases from 38.04\% to 14.72\%
across the sweep, while $\tau=0.30$ gives the lowest loss and PPL.

\subsection{Comparison with Periodic Updates}

The model-level periodic baseline applies full updates to all eligible
matrices at fixed intervals. Muon-LoGo instead uses the current score of
each matrix. Table~\ref{tab:ablation-model-periodic} reports the periodic
baseline. Its best setting reaches validation loss 3.2250, while
Muon-LoGo reaches 3.2128.

\subsection{Why Spectral-Norm CV Is Used}

Spectral-norm CV is a low-cost monitoring statistic rather than a
complete description of cross-shard matrix structure. It compares the
largest singular values across shards but does not capture singular-vector
directions or smaller singular values. We use it as a practical proxy for
deciding when full orthogonalization is useful.
To test the choice of monitoring signal, we compare spectral-norm CV
with three alternatives under the same optimization setting.
Table~\ref{tab:monitor-signal-main} compares the four signals under the
same threshold. Spectral-norm CV reaches validation loss 3.2128 and PPL
24.85. The alternatives reach losses of 3.2533--3.2995 and PPL values of
25.88--27.10. Spectral-norm CV also has the lowest full-orthogonalization
communication at 340.58B elements and a 35.0\% size-weighted full-update
ratio. The alternatives communicate 366.18--371.78B elements and have
ratios of 37.6--38.2\%. The shared threshold allows each signal to
determine its own full-update ratio.

\subsection{Why Matrix-Specific Normalization Is Needed}

Baseline spectral-norm CV values differ across attention, MLP, router,
and expert matrices, so a raw threshold would conflate matrix scales.
Matrix-specific statistics compare each CV with the recent history of
the same matrix, giving the threshold a common interpretation.

\subsection{Selected Matrix Sizes}

Muon-LoGo's unweighted and size-weighted full-update ratios are 36.0\%
and 35.0\%. The one-percentage-point difference indicates that selected
and eligible matrices have similar average sizes.

\vspace{0.6\baselineskip}
\section{Conclusion}

Muon-LoGo selects blockwise or full orthogonalization for each matrix
using normalized spectral-norm CV. The analysis gives norm control,
endpoint recovery, and per-matrix convergence for adaptive geometry
switching. The reference-error bound requires explicit score calibration,
while the communication result covers independent matrix-level decisions.
Muon-LoGo improves over Blockwise Muon on both models. Its MoE validation
loss is within 0.5\% of Full Muon with 65.0\% less communication.

\newpage
\section{Limitations}

The experiments cover 0.4B dense and MoE models trained on English C4, due to limited computational resources.
Larger models, other data domains, and multiple seeds would broaden the evaluation.
The convergence theorem analyzes one parameter matrix and one update
decision per step. It does not cover a product space with independent
matrix-level decisions in the same step.
Lemma~\ref{lem:monitor-control} assumes pointwise score calibration for
its reference-error bound relative to Full Muon. The analysis does not derive this
calibration from spectral-norm CV. Future work will study product-space
convergence and weaker average calibration across optimizer steps.
The per-matrix convergence result and communication identity require no
calibration.

\section{Ethical Consideration}

Muon-LoGo is an optimization method for language model pretraining and
does not introduce a new dataset or release a trained model. However,
improved training methods can support both beneficial and harmful uses of
language models. Models trained with Muon-LoGo may inherit biases,
offensive content, privacy risks, and other limitations from their
training data. Our experiments use English C4 and therefore also inherit
the known risks of large-scale web data. Model developers should evaluate
the resulting models for their intended applications and apply appropriate
data filtering, safety evaluation, and deployment safeguards.

\bibliography{custom}

\clearpage

\appendix

\section{Additional Implementation Details}
\label{app:implementation-details}

\subsection{Systems and compute}

All 0.4B experiments use four NVIDIA A800 GPUs with four-way tensor
parallelism. The software stack uses PyTorch 2.10, CUDA 12.8, and
Megatron-LM \cite{shoeybi2019megatronlm}.
The A800 nodes have different CPU configurations, so their per-step
wall-clock measurements are not directly comparable.

\subsection{Post-backward communication execution}
\label{app:communication-execution}

Blockwise Muon, Muon-LoGo, and Full Muon share the same tensor-parallel
activation and gradient collectives.
They differ only after backward, when the optimizer orthogonalizes each
local momentum shard.

\paragraph{Blockwise Muon.}
Newton--Schulz is applied independently to each local shard.
The update requires no remote matrix block and is applied directly to
the local parameter shard.
Blockwise Muon adds no optimizer-induced matrix communication but omits
the aggregate Gram matrix used by full orthogonalization.

\paragraph{Muon-LoGo.}
Each rank estimates the spectral norm of every eligible matrix shard.
Two scalar all-reduces form the CV after warmup and produce the same
decision across the tensor-parallel group.
A blockwise update follows Blockwise Muon.
For a full update, distributed Newton--Schulz performs full
orthogonalization and returns the corresponding update shard.
One optimizer step may contain both update types.

\paragraph{Full Muon.}
Every eligible matrix follows the distributed path at every step.
Full Muon does not use the monitoring signal and performs full
orthogonalization for every matrix.

\paragraph{Communication accounting.}
The cumulative communication is the sum of the full-matrix element
counts over all full updates.
The resulting quantity equals
$\sum_{t,i}a_{t,i}q_i=\rho_T^{\mathrm{w}}TQ$ in
Proposition~\ref{prop:communication-accounting}.
It excludes scalar monitoring collectives.

\subsection{Spectral-norm estimation}

Each TP rank estimates the spectral norm of $G_{t,k}$ using the power
iteration in Section~\ref{sec:method}.
Each iteration uses one multiplication by $G_{t,k}$ and one by
$G_{t,k}^{\top}$.
With two iterations, this local computation is inexpensive relative to
Newton--Schulz orthogonalization.
Only the scalar estimate is synchronized.

\section{Hyperparameter Sweep Results}
\label{app:sweep-results}

We report two-stage sweeps for Blockwise Muon and AdamW.
All candidates share the token budget, cosine schedule, and evaluation protocol.

\subsection{Search protocol}

Configurations from similarly sized pretrained models define the initial
search ranges \cite{wen2025fantastic,vyas2025soap,jordan2024muon}.
The first stage searches a broad range of learning rates and weight
decay values.
The second stage refines the best first-stage setting.
We select the configuration with the lowest endpoint validation loss.

\subsection{Blockwise Muon}

Table~\ref{tab:sweep-blockwise-muon} reports eight configurations.
The reference configuration is
$\eta=1.5\times10^{-2}$, $\lambda=0.05$, and $\mu=0.98$.
Stage~1 evaluates six learning-rate and weight-decay combinations, and
Stage~2 evaluates two settings near the best Stage~1 result.
All candidates use $\mu=0.98$, Nesterov momentum, five Newton--Schulz steps, unit-RMS scaling, and 400 warmup steps.

\begin{table*}[t]
\centering
\setlength{\tabcolsep}{8pt}
\renewcommand{\arraystretch}{1.08}
\caption{Two-stage hyperparameter sweep for Muon on the 0.4B dense
model. All runs use the same matched budget. The selected configuration
(LR${}=2\times10^{-2}$, weight decay${}=0.1$) is highlighted in bold.}
\label{tab:sweep-blockwise-muon}
\begin{tabular}{lcccccc}
\toprule
Stage & Candidate & LR ($\eta$) & Weight Decay ($\lambda$) & $\mu$ & Val Loss $\downarrow$ & Val PPL $\downarrow$ \\
\midrule
\multicolumn{7}{c}{\textit{Stage 1, Broad Search}} \\
\midrule
& 1 & $1.0\times10^{-2}$ & 0.05 & 0.98 & $3.064$ & $21.41$ \\
& 2 & $1.0\times10^{-2}$ & 0.10 & 0.98 & $3.049$ & $21.09$ \\
& 3 & $1.5\times10^{-2}$ & 0.05 & 0.98 & $3.039$ & $20.88$ \\
& 4 & $1.5\times10^{-2}$ & 0.10 & 0.98 & $3.026$ & $20.61$ \\
& 5 & $2.0\times10^{-2}$ & 0.05 & 0.98 & $3.021$ & $20.51$ \\
& 6 & $\mathbf{2.0\times10^{-2}}$ & $\mathbf{0.10}$ & 0.98 & $\mathbf{3.01}$ & $\mathbf{20.29}$ \\
\midrule
\multicolumn{7}{c}{\textit{Stage 2, Refinement}} \\
\midrule
& 7 & $2.0\times10^{-2}$ & 0.08 & 0.98 & $3.018$ & $20.45$ \\
& 8 & $2.5\times10^{-2}$ & 0.10 & 0.98 & $3.024$ & $20.57$ \\
\bottomrule
\end{tabular}
\end{table*}

Candidate~6 has the lowest validation loss.
We use LR${}=2.0\times10^{-2}$ and weight decay${}=0.1$ for the
Blockwise Muon result in Table~\ref{tab:main-result-dense-model}.

\subsection{AdamW}

Table~\ref{tab:sweep-adamw} reports eight AdamW configurations on the
0.4B dense model.
Stage~1 evaluates learning rates in $\{1.0,1.5,2.0\}\times10^{-2}$ with weight decay in $\{0.01,0.05\}$.
Stage~2 evaluates a lower learning rate with the same two weight decay values.
All candidates use $\beta_1=0.9$, $\beta_2=0.98$, $\epsilon=10^{-8}$, decoupled weight decay, and 400 warmup steps.

\begin{table*}[t]
\centering
\setlength{\tabcolsep}{7pt}
\renewcommand{\arraystretch}{1.08}
\caption{Two-stage hyperparameter sweep for AdamW on the 0.4B dense model. The selected configuration
(LR${}=1.0\times10^{-2}$, weight decay${}=0.05$) is highlighted in bold.}
\label{tab:sweep-adamw}
\begin{tabular}{lcccccc}
\toprule
Stage & Candidate & LR ($\eta$) & Weight Decay ($\lambda$) & $\beta_1$ / $\beta_2$ & Val Loss $\downarrow$ & Val PPL $\downarrow$ \\
\midrule
\multicolumn{7}{c}{\textit{Stage 1, Broad Search}} \\
\midrule
& 1 & $1.0\times10^{-2}$ & 0.01 & 0.9 / 0.98 & 3.305 & 27.25 \\
& 2 & $\mathbf{1.0\times10^{-2}}$ & $\mathbf{0.05}$ & 0.9 / 0.98 & $\mathbf{3.22}$ & $\mathbf{25.03}$ \\
& 3 & $1.5\times10^{-2}$ & 0.01 & 0.9 / 0.98 & 3.251 & 25.82 \\
& 4 & $1.5\times10^{-2}$ & 0.05 & 0.9 / 0.98 & 3.257 & 25.97 \\
& 5 & $2.0\times10^{-2}$ & 0.01 & 0.9 / 0.98 & 3.264 & 26.15 \\
& 6 & $2.0\times10^{-2}$ & 0.05 & 0.9 / 0.98 & 3.420 & 30.57 \\
\midrule
\multicolumn{7}{c}{\textit{Stage 2, Refinement}} \\
\midrule
& 7 & $0.5\times10^{-2}$ & 0.01 & 0.9 / 0.98 & 3.320 & 27.66 \\
& 8 & $0.5\times10^{-2}$ & 0.05 & 0.9 / 0.98 & 3.314 & 27.49 \\
\bottomrule
\end{tabular}
\end{table*}

Candidate~2 has the lowest validation loss.
We use LR${}=1.0\times10^{-2}$ and weight decay${}=0.05$
for the AdamW result in Table~\ref{tab:main-result-dense-model}.

\subsection{Additional 0.4B MoE-Model Sweeps}

Tables~\ref{tab:sweep-blockwise-muon-dense-additional}
and~\ref{tab:sweep-adamw-moe} report additional 0.4B MoE sweeps.

\begin{table*}[t]
\centering
\setlength{\tabcolsep}{8pt}
\renewcommand{\arraystretch}{1.08}
\caption{Additional two-stage hyperparameter sweep for Muon on
the 0.4B MoE model. All runs use the same training budget and
evaluation protocol.}
\label{tab:sweep-blockwise-muon-dense-additional}
\begin{tabular}{lcccccc}
\toprule
Stage & Candidate & LR ($\eta$) & Weight Decay ($\lambda$) & $\mu$ & Val Loss $\downarrow$ & Val PPL $\downarrow$ \\
\midrule
\multicolumn{7}{c}{\textit{Stage 1, Broad Search}} \\
\midrule
& 1 & $1.0\times10^{-3}$ & 0.15 & 0.98 & 3.6309 & 37.75 \\
& 2 & $2.0\times10^{-3}$ & 0.15 & 0.98 & 3.4059 & 30.14 \\
& 3 & $5.0\times10^{-3}$ & 0.15 & 0.98 & 3.2709 & 26.34 \\
& 4 & $1.0\times10^{-2}$ & 0.15 & 0.98 & 3.2169 & 24.95 \\
& 5 & $1.5\times10^{-2}$ & 0.15 & 0.98 & 3.1979 & 24.48 \\
& 6 & $2.0\times10^{-2}$ & 0.15 & 0.98 & 3.1989 & 24.51 \\
& 7 & $3.0\times10^{-2}$ & 0.15 & 0.98 & 3.2039 & 24.63 \\
& 8 & $5.0\times10^{-2}$ & 0.15 & 0.98 & 3.2149 & 24.90 \\
\midrule
\multicolumn{7}{c}{\textit{Stage 2, Refinement}} \\
\midrule
& 9 & $\mathbf{1.5\times10^{-2}}$ & $\mathbf{0.05}$ & 0.98 & $\mathbf{3.1969}$ & $\mathbf{24.46}$ \\
& 10 & $1.5\times10^{-2}$ & 0.10 & 0.98 & 3.2069 & 24.70 \\
& 11 & $1.5\times10^{-2}$ & 0.15 & 0.98 & 3.2109 & 24.80 \\
& 12 & $1.5\times10^{-2}$ & 0.25 & 0.98 & 3.1999 & 24.53 \\
\bottomrule
\end{tabular}
\end{table*}

\begin{table*}[t]
\centering
\setlength{\tabcolsep}{7pt}
\renewcommand{\arraystretch}{1.08}
\caption{Additional two-stage hyperparameter sweep for AdamW on the
0.4B MoE model. All runs use the same training budget and evaluation
protocol.}
\label{tab:sweep-adamw-moe}
\begin{tabular}{lcccccc}
\toprule
Stage & Candidate & LR ($\eta$) & Weight Decay ($\lambda$) & $\beta_1$ / $\beta_2$ & Val Loss $\downarrow$ & Val PPL $\downarrow$ \\
\midrule
\multicolumn{7}{c}{\textit{Stage 1, Broad Search}} \\
\midrule
& 1 & \textbf{0.2e-2} & \textbf{0.1} & 0.9 / 0.98 & \textbf{3.320} & \textbf{27.67} \\
& 2 & 0.2e-2 & 0.2 & 0.9 / 0.98 & 3.323 & 27.74 \\
& 3 & 0.5e-2 & 0.1 & 0.9 / 0.98 & 3.357 & 28.70 \\
& 4 & 0.5e-2 & 0.2 & 0.9 / 0.98 & 3.327 & 27.85 \\
& 5 & 0.8e-2 & 0.1 & 0.9 / 0.98 & 3.413 & 30.36 \\
& 6 & 0.8e-2 & 0.2 & 0.9 / 0.98 & 3.397 & 29.87 \\
\midrule
\multicolumn{7}{c}{\textit{Stage 2, Refinement}} \\
\midrule
& 7 & 0.1e-2 & 0.1 & 0.9 / 0.98 & 3.329 & 27.91 \\
& 8 & 0.1e-2 & 0.2 & 0.9 / 0.98 & 3.315 & 27.52 \\
\bottomrule
\end{tabular}
\end{table*}

\section{Additional Ablation Results}
\label{app:ablation-results}

Table~\ref{tab:monitor-signal-main} reports the monitoring-signal
comparison in the main text.
The remaining ablations examine a model-level periodic baseline and the
normalized-score threshold.
The threshold sweep studies the trade-off between full-update frequency
and validation quality.

\subsection{Model-Level Periodic Updates}

The model-level periodic baseline applies one schedule to the entire
model.
Every $K$ steps, the method applies a full update to all eligible
matrices.
All other steps use blockwise updates.
The full-update frequency is therefore determined by the interval and is
approximately $1/K$.
Table~\ref{tab:ablation-model-periodic} reports four intervals on the
same 0.4B MoE model.

\begin{table*}[t]
\centering
\setlength{\tabcolsep}{7pt}
\renewcommand{\arraystretch}{1.08}
\caption{Model-level periodic-update sweep on the 0.4B MoE model after 4,000 optimizer steps.}
\label{tab:ablation-model-periodic}
\begin{tabular}{cccc}
\toprule
Interval $K$ &  Val Loss $\downarrow$ & Val PPL $\downarrow$ & Full-Update Frequency $\downarrow$ \\
\midrule
2 & 3.2259 & 25.18 & 50.00\%\\
3 & \textbf{3.2250} & \textbf{25.15} & 33.30\% \\
4 & 3.2328 & 25.35 & 25.00\%\\
6 & 3.2261 & 25.18 & 16.65\%\\
\bottomrule
\end{tabular}
\end{table*}

The $K=2$, $K=3$, and $K=6$ runs reach validation losses between 3.2250
and 3.2261, although their full-update frequencies range from 16.65\% to
50.0\%.
Interval $K=6$ therefore reaches similar endpoint quality to $K=2$ with
about one third of the full-update steps.
The $K=4$ run reaches a higher validation loss of 3.2328 and does not
follow an ordering based on communication frequency.
The sweep therefore provides no consistent relation between a shorter
interval and a better endpoint.

The model-level schedule uses the same decision for matrices with
different update statistics.
A matrix with balanced shards receives a full update whenever the global
schedule triggers, while a matrix with an unusual shard imbalance must
wait for the next scheduled step.
Muon-LoGo removes the shared timing constraint and evaluates each matrix
using its own recent statistics.

\subsection{Normalized-Score Threshold}

The normalized-score threshold controls how much evidence is required
before Muon-LoGo applies a full update.
A lower threshold accepts smaller deviations from the recent
matrix-specific statistics and therefore triggers more full updates.
A higher threshold triggers full updates only for larger deviations.
To measure the sensitivity to this choice,
Table~\ref{tab:ablation-z-threshold} varies $\tau$ on the same 0.4B MoE
model.
All runs use spectral-norm CV, the same learning rate, the same
normalization, and the same training budget.

\begin{table*}[t]
\centering
\setlength{\tabcolsep}{7pt}
\renewcommand{\arraystretch}{1.08}
\caption{Muon-LoGo normalized-score threshold sweep on the 0.4B MoE
model. The unweighted full-update ratio is the fraction of matrix-level
decisions assigned a full update.}
\label{tab:ablation-z-threshold}
\begin{tabular}{ccccc}
\toprule
Threshold $\tau$ & Val Loss $\downarrow$ & Val PPL $\downarrow$ & Unweighted Full-Update Ratio $\downarrow$ \\
\midrule
0.15 & 3.2293 & 25.26 & 38.04\% \\
0.30 & \textbf{3.2128} & \textbf{24.85} & 35.98\% \\
0.50 & 3.2142 & 24.88 & 26.86\% \\
0.80 & 3.2157 & 24.92 & 19.15\% \\
1.00 & 3.2166 & 24.94 & 14.72\% \\
\bottomrule
\end{tabular}
\end{table*}

The unweighted full-update ratio decreases steadily as $\tau$ increases.
Raising $\tau$ from 0.30 to 1.00 reduces the ratio from 35.98\% to
14.72\%.
The corresponding validation loss changes from 3.2128 to 3.2166, and
PPL changes from 24.85 to 24.94.
The settings $\tau=0.50$ and $\tau=0.80$ provide intermediate
communication levels with gradual changes in endpoint quality.

The lowest threshold does not give the best result.
At $\tau=0.15$, Muon-LoGo performs more full updates than at
$\tau=0.30$ but reaches validation loss 3.2293 and PPL 25.26.
The sweep does not support a simple explanation based on increasing the
number of full updates.
The threshold controls communication directly, but validation quality is
best at an intermediate value in the tested range.

\subsection{Monitoring Signals}

The comparison in Table~\ref{tab:monitor-signal-main} uses four local shard
statistics. For a scalar shard statistic $x_{t,k}$, define
\begin{equation}
\operatorname{CV}_k(x_{t,k})
=
\frac{\operatorname{std}_k(x_{t,k})}
{\operatorname{mean}_k(x_{t,k})+\epsilon}.
\end{equation}
Energy CV uses the squared Frobenius norm:
\begin{equation}
c_t^{\mathrm{energy}}
=
\operatorname{CV}_k\!\left(\|G_{t,k}\|_{\mathrm{F}}^2\right).
\end{equation}
Stable-rank CV uses the stable-rank estimate:
\begin{equation}
c_t^{\mathrm{sr}}
=
\operatorname{CV}_k\!\left(
\frac{\|G_{t,k}\|_{\mathrm{F}}^2}
{s_{t,k}^2+\epsilon}
\right).
\end{equation}
Spectral-norm CV is $c_t^{\mathrm{spec}}$ from
Section~\ref{sec:method}. For directional Gram CV, let
$V=[v_1,\ldots,v_q]$ contain fixed normalized probe vectors and define
$e_{t,k,j}=\|G_{t,k}v_j\|_2^2$. The signal averages the cross-shard CV
over the probes:
\begin{equation}
c_t^{\mathrm{dir}}
=
\frac{1}{q}\sum_{j=1}^{q}
\operatorname{CV}_k(e_{t,k,j}).
\end{equation}
All four signals use the same matrix-specific normalization, threshold,
and decision rule.

\section{Auxiliary Theoretical Results}

\DetailedConvergenceResults

\begin{lemma}[Norm control of orthogonalized updates]
\label{lem:norm-control}
Let $X=[X_1,\ldots,X_K]$ be the canonical horizontal partition defined
in Section~\ref{sec:background} and define
\begin{equation}
\|X\|_{\mathrm{F}}=\|X\|_{\mathrm{op}},
\qquad
\|X\|_{\mathrm{B}}=\max_{1\le b\le K}\|X_b\|_{\mathrm{op}}.
\end{equation}
Then
\begin{equation}
\begin{aligned}
\|X\|_{\mathrm{B}}
&\le
\|X\|_{\mathrm{F}}
\le
\left(\sum_{b=1}^{K}\|X_b\|_{\mathrm{op}}^2\right)^{1/2}
\\[-2pt]
&\quad\le
\sqrt{K}\,\|X\|_{\mathrm{B}}.
\end{aligned}
\end{equation}
For
\begin{equation}
\begin{aligned}
U_t^{\mathrm{F}}
&=\operatorname{Polar}(M_t),\\
U_t^{\mathrm{B}}
&=[
\operatorname{Polar}(M_{t,1}),\ldots,\\
&\hspace{3.3em}
\operatorname{Polar}(M_{t,K})
].
\end{aligned}
\end{equation}
one has
\begin{equation}
\begin{aligned}
\|U_t^{\mathrm{F}}\|_{\mathrm{F}}
&\le1,
&
\|U_t^{\mathrm{F}}\|_{\mathrm{B}}
&\le1,\\
\|U_t^{\mathrm{B}}\|_{\mathrm{B}}
&\le1,
&
\|U_t^{\mathrm{B}}\|_{\mathrm{F}}
&\le\sqrt K.
\end{aligned}
\end{equation}
Consequently,
\begin{equation}
\omega_t
:=
\|U_t^{\mathrm{B}}-U_t^{\mathrm{F}}\|_{\mathrm{F}}
\le
1+\sqrt K.
\end{equation}
\end{lemma}

\begin{lemma}[Conditional approximation error under score calibration]
\label{lem:monitor-control}
Let $a_t=\mathbf{1}[z_t\ge z_{\mathrm{thr}}]$ and let
$j_t=\mathrm{F}$ when $a_t=1$ and $j_t=\mathrm{B}$ otherwise.
Assume an externally specified nondecreasing calibration function
$\phi:\mathbb{R}\to\mathbb{R}_{+}$ satisfies
$\omega_t\le\phi(z_t)$ for every $t$, and suppose
\begin{equation}
\|\widehat U_t-U_t^{j_t}\|_{j_t}\le\delta_t,
\qquad
0\le\delta_t<1.
\end{equation}
Then
\begin{equation}
\|\widehat U_t-U_t^{\mathrm{F}}\|_{\mathrm{F}}
\le
a_t\delta_t
+
(1-a_t)
\bigl(\phi(z_{\mathrm{thr}})+\sqrt K\,\delta_t\bigr).
\end{equation}
For any nonnegative weights $\{\eta_t\}_{t=0}^{T-1}$,
\begin{equation}
\begin{aligned}
\sum_{t<T}\eta_t
\|\widehat U_t-U_t^{\mathrm{F}}\|_{\mathrm{F}}
&\le
\phi(z_{\mathrm{thr}})
\sum_{t<T}\eta_t(1-a_t)
\\
&\quad+
\sum_{t<T}\eta_t\delta_t\\[-2pt]
&\qquad\cdot
\bigl(a_t+\sqrt K(1-a_t)\bigr).
\end{aligned}
\end{equation}
\end{lemma}

\begin{corollary}[Reference error relative to Full Muon]
\label{cor:muon-logo-reference-error}
Define
\begin{equation}
\omega_t
=
\|U_t^{\mathrm{B}}-U_t^{\mathrm{F}}\|_{\mathrm{op}},
\end{equation}
and
\begin{equation}
\begin{aligned}
\epsilon_t^{\mathrm{F}}
&=
\|\widehat U_t^{\mathrm{F}}-U_t^{\mathrm{F}}\|_{\mathrm{op}},
\\
\epsilon_t^{\mathrm{B}}
&=
\|\widehat U_t^{\mathrm{B}}-U_t^{\mathrm{B}}\|_{\mathrm{op}}.
\end{aligned}
\end{equation}
The selected direction satisfies
\begin{equation}
\begin{aligned}
e_t
&:=
\|\widehat U_t-U_t^{\mathrm{F}}\|_{\mathrm{op}}
\\
&\le
a_t\epsilon_t^{\mathrm{F}}
+
(1-a_t)
\left(\omega_t+\epsilon_t^{\mathrm{B}}\right).
\end{aligned}
\label{eq:update-gap}
\end{equation}
Suppose
$\|M_t-\nabla f(X_t)\|_*\le\nu_t^{\mathrm{op}}$ almost surely.
If $e_t<1$ for every $t$, then
\begin{equation}
\begin{aligned}
&\min_{0\le t<T}\|\nabla f(X_t)\|_{*}
\\
&\quad\le
\frac{
\displaystyle
\Delta_0
+
2\sum_{t<T}\eta_t\nu_t^{\mathrm{op}}
+
\frac{L_{\mathrm{op}}}{2}
\sum_{t<T}\eta_t^2(1+e_t)^2
}{
\displaystyle
\sum_{t<T}\eta_t(1-e_t)
}.
\end{aligned}
\label{eq:full-reference-general}
\end{equation}
In particular, suppose $\epsilon_t^{\mathrm{F}}=0$ and
\begin{equation}
\omega_t+\epsilon_t^{\mathrm{B}}
\le\bar\omega<1
\end{equation}
on every blockwise step. With a constant stepsize $\eta$ and
$\nu_t^{\mathrm{op}}=0$,
\begin{equation}
\begin{aligned}
\min_{0\le t<T}\|\nabla f(X_t)\|_{*}
&\le
\frac{\Delta_0}{
\eta T[1-(1-r_T)\bar\omega]
}
\\
&\quad+
\frac{L_{\mathrm{op}}\eta}{2}\\[-2pt]
&\qquad\times
\tfrac{
1+(1-r_T)(2\bar\omega+\bar\omega^2)
}{
1-(1-r_T)\bar\omega
}.
\end{aligned}
\label{eq:full-reference-constant}
\end{equation}
Choosing
\begin{equation}
\eta^\star
=
\sqrt{
\frac{
2\Delta_0
}{
L_{\mathrm{op}}T
[1+(1-r_T)(2\bar\omega+\bar\omega^2)]
}
}
\end{equation}
gives
\begin{equation}
\begin{aligned}
&\min_{0\le t<T}\|\nabla f(X_t)\|_{*}
\\
&\quad\le
\sqrt{\frac{2L_{\mathrm{op}}\Delta_0}{T}}
\,
\frac{
\sqrt{1+(1-r_T)(2\bar\omega+\bar\omega^2)}
}{
1-(1-r_T)\bar\omega
}.
\end{aligned}
\label{eq:full-reference-rate}
\end{equation}
The multiplicative factor is non-increasing in $r_T$ and equals one
when $r_T=1$ or $\bar\omega=0$.
\end{corollary}

\begin{proposition}[Threshold and communication trade-off]
\label{prop:threshold-tradeoff}
Fix one realized score sequence $\{z_t\}_{t=0}^{T-1}$ and nonnegative
weights $\{\eta_t\}_{t=0}^{T-1}$. Define
\begin{equation}
r_T(\tau)
=
\frac1T\sum_{t<T}\mathbf{1}[z_t\ge\tau].
\end{equation}
Then $r_T(\tau)$ is non-increasing in $\tau$. If the calibration
function $\phi$ in Lemma~\ref{lem:monitor-control} is nonnegative and
nondecreasing, then
\begin{equation}
R_T(\tau)
=
\phi(\tau)\sum_{t<T}\eta_t\mathbf{1}[z_t<\tau]
\end{equation}
is nondecreasing in $\tau$. Thus lowering the threshold weakly
increases full updates and communication while weakly decreasing
the calibrated blockwise-error bound for the fixed score sequence.
\end{proposition}

\section{Proofs}

\subsection{Proof of Proposition~\ref{prop:special-cases}}

\begin{proof}
For a fixed finite post-warmup horizon, every realized score $z_t$ is
finite. If
$z_{\mathrm{thr}}>\max_{t<T}z_t$, then
$\mathbf{1}[z_t\ge z_{\mathrm{thr}}]=0$ for every $t$, so every
selected update is blockwise. If
$z_{\mathrm{thr}}\le\min_{t<T}z_t$, the indicator is one for every
$t$, so every post-warmup selected update is full. Taking the threshold to
$+\infty$ or $-\infty$ gives the two limiting post-warmup update rules.
If the warmup length is zero, the latter is Full Muon on the entire
trajectory.
\end{proof}

\subsection{Proof of Lemma~\ref{lem:norm-control}}

\begin{proof}
Let $E_b$ be the canonical column embedding for block $b$. Since
$X_b=XE_b$ and $\|E_b\|_{\mathrm{op}}=1$,
\begin{equation}
\|X_b\|_{\mathrm{op}}
\le
\|X\|_{\mathrm{op}}\|E_b\|_{\mathrm{op}}
=
\|X\|_{\mathrm{op}}.
\end{equation}
Taking the maximum over $b$ proves
$\|X\|_{\mathrm{B}}\le\|X\|_{\mathrm{F}}$. Because the partition is
horizontal,
\begin{equation}
XX^\top=\sum_{b=1}^{K}X_bX_b^\top.
\end{equation}
The triangle inequality for the operator norm therefore gives
\begin{equation}
\begin{aligned}
\|X\|_{\mathrm{F}}^2
&=\|XX^\top\|_{\mathrm{op}}\\
&\le
\sum_{b=1}^{K}\|X_bX_b^\top\|_{\mathrm{op}}
=
\sum_{b=1}^{K}\|X_b\|_{\mathrm{op}}^2\\
&\le K\|X\|_{\mathrm{B}}^2.
\end{aligned}
\end{equation}
Taking square roots proves the norm chain.

If $A=P\Sigma Q^\top$ is a compact singular value decomposition, then
$\operatorname{Polar}(A)=PQ^\top$ on the support of $A$ and is a
partial isometry. Hence its operator norm is at most one. Applying this
fact to $M_t$ gives
$\|U_t^{\mathrm{F}}\|_{\mathrm{F}}\le1$, and the first norm inequality
also gives $\|U_t^{\mathrm{F}}\|_{\mathrm{B}}\le1$. Applying it
separately to every block gives
$\|U_t^{\mathrm{B}}\|_{\mathrm{B}}\le1$; the last norm inequality then
gives $\|U_t^{\mathrm{B}}\|_{\mathrm{F}}\le\sqrt K$. Finally,
\begin{equation}
\omega_t
\le
\|U_t^{\mathrm{B}}\|_{\mathrm{F}}
+
\|U_t^{\mathrm{F}}\|_{\mathrm{F}}
\le
\sqrt K+1.
\end{equation}
\end{proof}

\subsection{Proof of Lemma~\ref{lem:monitor-control}}

\begin{proof}
If $a_t=1$, then $j_t=\mathrm{F}$ and the inexact-oracle condition
directly gives
\begin{equation}
\|\widehat U_t-U_t^{\mathrm{F}}\|_{\mathrm{F}}
\le\delta_t.
\end{equation}
If $a_t=0$, then $j_t=\mathrm{B}$ and $z_t<z_{\mathrm{thr}}$.
Monotonicity of $\phi$ yields
\begin{equation}
\omega_t
\le
\phi(z_t)
\le
\phi(z_{\mathrm{thr}}).
\end{equation}
The norm comparison in Lemma~\ref{lem:norm-control}, applied to
$\widehat U_t-U_t^{\mathrm{B}}$, gives
\begin{equation}
\|\widehat U_t-U_t^{\mathrm{B}}\|_{\mathrm{F}}
\le
\sqrt K
\|\widehat U_t-U_t^{\mathrm{B}}\|_{\mathrm{B}}
\le
\sqrt K\,\delta_t.
\end{equation}
Thus, on a blockwise step, the triangle inequality gives
\begin{equation}
\|\widehat U_t-U_t^{\mathrm{F}}\|_{\mathrm{F}}
\le
\sqrt K\,\delta_t+\phi(z_{\mathrm{thr}}).
\end{equation}
Combining the two cases proves the pointwise bound. Multiplying it by
$\eta_t\ge0$ and summing over $t$ proves the cumulative bound.
\end{proof}

\subsection{Proof of Theorem~\ref{thm:muon-logo-convergence}}

\begin{proof}
Write $g_t=\nabla f(X_t)$ and $q_t=M_t-g_t$. The exact
linear-minimization oracle satisfies
\begin{equation}
\langle M_t,U_t^{j_t}\rangle
=
\|M_t\|_{j_t,*},
\qquad
\|U_t^{j_t}\|_{j_t}\le1.
\end{equation}
By duality and the inexact-oracle assumption,
\begin{equation}
\begin{aligned}
\langle M_t,\widehat U_t\rangle
&=
\langle M_t,U_t^{j_t}\rangle
+
\langle M_t,\widehat U_t-U_t^{j_t}\rangle\\
&\ge
\|M_t\|_{j_t,*}
-
\|M_t\|_{j_t,*}
\|\widehat U_t-U_t^{j_t}\|_{j_t}\\
&\ge
(1-\delta_t)\|M_t\|_{j_t,*}.
\end{aligned}
\end{equation}
The triangle inequality also yields
$\|\widehat U_t\|_{j_t}\le1+\delta_t$. Therefore,
\begin{equation}
\begin{aligned}
\langle g_t,\widehat U_t\rangle
&=
\langle M_t,\widehat U_t\rangle
-
\langle q_t,\widehat U_t\rangle\\
&\ge
(1-\delta_t)\|M_t\|_{j_t,*}
-
(1+\delta_t)\|q_t\|_{j_t,*}.
\end{aligned}
\end{equation}
The reverse triangle inequality and
$\|q_t\|_{j_t,*}\le\nu_t$ imply
\begin{equation}
\|M_t\|_{j_t,*}
\ge
\|g_t\|_{j_t,*}-\nu_t.
\end{equation}
Substitution gives
\begin{equation}
\begin{aligned}
\langle g_t,\widehat U_t\rangle
&\ge
(1-\delta_t)\|g_t\|_{j_t,*}
\\[-2pt]
&\quad-
\bigl[(1-\delta_t)+(1+\delta_t)\bigr]\nu_t\\
&=
(1-\delta_t)\|g_t\|_{j_t,*}-2\nu_t\\
&\ge
(1-\delta_t)\|g_t\|_*-2\nu_t,
\end{aligned}
\end{equation}
where the final inequality uses
$\|g_t\|_*\le\|g_t\|_{j_t,*}$.

Apply $L_{j_t}$-smoothness with
$Y=X_t-\eta_t\widehat U_t$.
\begin{equation}
\begin{aligned}
f(X_{t+1})
&\le
f(X_t)
-
\eta_t\langle g_t,\widehat U_t\rangle
+
\frac{L_t\eta_t^2}{2}
\|\widehat U_t\|_{j_t}^2\\
&\le
f(X_t)
-
\eta_t(1-\delta_t)\|g_t\|_*
+
2\eta_t\nu_t\\
&\quad+
\frac{L_t\eta_t^2}{2}(1+\delta_t)^2.
\end{aligned}
\end{equation}
Rearranging, summing from $t=0$ to $T-1$, and using
$f(X_T)\ge f_\star$ proves
Equation~\eqref{eq:mixed-cumulative}. Since all weights
$\eta_t(1-\delta_t)$ are positive, their weighted average dominates
the minimum, which proves Equation~\eqref{eq:mixed-min}.

For the asymptotic statement, let
$S_T=\sum_{t<T}\eta_t$,
$\bar\delta=\sup_t\delta_t<1$, and
$L_{\max}=\max\{L_{\mathrm{op}},L_{\mathrm{B}}\}$.
Suppose for contradiction that
$\liminf_{t\to\infty}\|\nabla f(X_t)\|_*>0$.
Then there is a constant $\gamma>0$ and an index $T_0$ such that
$\|\nabla f(X_t)\|_*\ge\gamma$ for every $t\ge T_0$.
The left-hand side of Equation~\eqref{eq:mixed-cumulative} therefore
satisfies
\begin{equation}
\begin{aligned}
&\sum_{t<T}\eta_t(1-\delta_t)
\|\nabla f(X_t)\|_*\\
&\quad\ge
\gamma(1-\bar\delta)
\sum_{t=T_0}^{T-1}\eta_t.
\end{aligned}
\end{equation}
Because $\sum_t\eta_t=\infty$, the tail sum diverges.
On the other hand, Equation~\eqref{eq:mixed-cumulative} and
$1+\delta_t\le2$ give
\begin{equation}
\begin{aligned}
&\frac{1}{S_T}
\sum_{t<T}\eta_t(1-\delta_t)
\|\nabla f(X_t)\|_*\\
&\quad\le
\frac{\Delta_0}{S_T}
+
2\frac{\sum_{t<T}\eta_t\nu_t}{S_T}\\
&\qquad+
\frac{2L_{\max}\sum_{t=0}^{\infty}\eta_t^2}{S_T}
\longrightarrow0.
\end{aligned}
\end{equation}
The tail lower bound divided by $S_T$ converges to
$\gamma(1-\bar\delta)>0$, a contradiction.
Hence
$\liminf_{t\to\infty}\|\nabla f(X_t)\|_*=0$.
\end{proof}

\subsection{Proof of Corollary~\ref{cor:muon-logo-interpolation}}

\begin{proof}
Set $\nu_t=\delta_t=0$ in
Equation~\eqref{eq:mixed-min} and define
\begin{equation}
S
=
\frac{N_{\mathrm{full}}}{L_{\mathrm{op}}}
+
\frac{N_{\mathrm{block}}}{L_{\mathrm{B}}}.
\end{equation}
With the stated geometry-dependent stepsizes,
\begin{equation}
\sum_{t<T}\eta_t=cS,
\qquad
\sum_{t<T}L_t\eta_t^2=c^2S.
\end{equation}
Equation~\eqref{eq:mixed-min} therefore becomes
\begin{equation}
\min_{t<T}\|\nabla f(X_t)\|_*
\le
\frac{\Delta_0}{cS}+\frac c2.
\end{equation}
Differentiating the right-hand side over $c>0$ gives the unique
minimizer $c^\star=\sqrt{2\Delta_0/S}$ and the value
$\sqrt{2\Delta_0/S}$, proving
Equation~\eqref{eq:interpolation-counts}. Because
\begin{equation}
S
=
T\left(
\frac{r_T}{L_{\mathrm{op}}}
+
\frac{1-r_T}{L_{\mathrm{B}}}
\right)
=
\frac{T}{L_{\mathrm{eff}}(r_T)},
\end{equation}
Equation~\eqref{eq:effective-smoothness} follows. Setting $r_T$ to
one or zero gives the two endpoint guarantees. Finally, when
$L_{\mathrm{op}}\le L_{\mathrm{B}}$, the denominator in the definition
of $L_{\mathrm{eff}}(r_T)$ is nondecreasing in $r_T$, so
$L_{\mathrm{eff}}(r_T)$ is non-increasing.
\end{proof}

\subsection{Proof of Corollary~\ref{cor:muon-logo-reference-error}}

\begin{proof}
On a full step, $\widehat U_t=\widehat U_t^{\mathrm{F}}$, so
$e_t=\epsilon_t^{\mathrm{F}}$. On a blockwise step,
$\widehat U_t=\widehat U_t^{\mathrm{B}}$, and the triangle inequality
gives
\begin{equation}
e_t
\le
\|\widehat U_t^{\mathrm{B}}-U_t^{\mathrm{B}}\|_{\mathrm{op}}
+
\|U_t^{\mathrm{B}}-U_t^{\mathrm{F}}\|_{\mathrm{op}}
=
\epsilon_t^{\mathrm{B}}+\omega_t.
\end{equation}
Equation~\eqref{eq:update-gap} follows.

Now use the full operator geometry at every step and view
$\widehat U_t$ as an $e_t$-inexact version of the full oracle
$U_t^{\mathrm{F}}$. The assumptions $e_t<1$ and
$\|M_t-\nabla f(X_t)\|_*\le\nu_t^{\mathrm{op}}$ match those of
Theorem~\ref{thm:muon-logo-convergence} with
$j_t=\mathrm{F}$ and $\delta_t=e_t$. Equation~\eqref{eq:mixed-min}
then becomes Equation~\eqref{eq:full-reference-general}.

Under the remaining assumptions, $e_t=0$ on each full step and
$e_t\le\bar\omega$ on each blockwise step. Let
$N_{\mathrm{full}}=r_TT$ and
$N_{\mathrm{block}}=(1-r_T)T$. Then
\begin{equation}
\begin{aligned}
\sum_{t<T}(1-e_t)
&\ge
N_{\mathrm{full}}
+
(1-\bar\omega)N_{\mathrm{block}}
\\[-2pt]
&=
T\bigl[1-(1-r_T)\bar\omega\bigr],
\end{aligned}
\end{equation}
and
\begin{equation}
\begin{aligned}
\sum_{t<T}(1+e_t)^2
&\le
N_{\mathrm{full}}
+
(1+\bar\omega)^2N_{\mathrm{block}}\\
&=
T\bigl[1+(1-r_T)(2\bar\omega+\bar\omega^2)\bigr].
\end{aligned}
\end{equation}
Substituting these inequalities and the constant stepsize into
Equation~\eqref{eq:full-reference-general} proves
Equation~\eqref{eq:full-reference-constant}.

Write
\begin{equation}
D=1-(1-r_T)\bar\omega,
\quad
N=1+(1-r_T)(2\bar\omega+\bar\omega^2).
\end{equation}
The right-hand side of
Equation~\eqref{eq:full-reference-constant} is
$\Delta_0/(\eta TD)+(L_{\mathrm{op}}N/(2D))\eta$.
Its minimizer is
$\eta^\star=\sqrt{2\Delta_0/(L_{\mathrm{op}}TN)}$, and substitution
gives Equation~\eqref{eq:full-reference-rate}. To prove monotonicity,
set $x=1-r_T$. The logarithmic derivative of
$\sqrt{1+x(2\bar\omega+\bar\omega^2)}/(1-x\bar\omega)$ with respect to
$x$ is
\begin{equation}
\frac{2\bar\omega+\bar\omega^2}
{2[1+x(2\bar\omega+\bar\omega^2)]}
+
\frac{\bar\omega}{1-x\bar\omega}
\ge0.
\end{equation}
The factor is therefore nondecreasing in $x$ and non-increasing in
$r_T$. The factor equals one when $r_T=1$ or $\bar\omega=0$.
\end{proof}

\subsection{Proof of Proposition~\ref{prop:threshold-tradeoff}}

\begin{proof}
For $\tau_1\le\tau_2$ and every $t$,
\begin{equation}
\mathbf{1}[z_t\ge\tau_1]
\ge
\mathbf{1}[z_t\ge\tau_2].
\end{equation}
Summing and dividing by $T$ proves
$r_T(\tau_1)\ge r_T(\tau_2)$. Equivalently,
$\mathbf{1}[z_t<\tau_1]\le\mathbf{1}[z_t<\tau_2]$, so
\begin{equation}
S_T(\tau)
:=
\sum_{t<T}\eta_t\mathbf{1}[z_t<\tau]
\end{equation}
is nondecreasing. The calibration function $\phi$ is also
nonnegative and nondecreasing. Hence, for $\tau_1\le\tau_2$,
\begin{equation}
\phi(\tau_1)S_T(\tau_1)
\le
\phi(\tau_2)S_T(\tau_1)
\le
\phi(\tau_2)S_T(\tau_2),
\end{equation}
which proves that $R_T(\tau)$ is nondecreasing. A lower threshold
therefore selects a superset of full updates while reducing the
calibrated blockwise-error term.
\end{proof}

\subsection{Proof of Proposition~\ref{prop:communication-accounting}}

\begin{proof}
At optimizer step $t$, matrix $i$ contributes $q_i$ logical full-tensor
elements if $a_{t,i}=1$ and contributes zero otherwise. Summing these
contributions over matrices and optimizer steps gives
\begin{equation}
\mathrm{Comm}_{\mathrm{full}}(T)
=
\sum_{t<T}\sum_{i=1}^{J}a_{t,i}q_i.
\end{equation}
By the definition of $\rho_T^{\mathrm{w}}$, the same sum equals
$\rho_T^{\mathrm{w}}TQ$. Dividing by $T$ gives the average
communication per optimizer step. Since $0\le a_{t,i}\le1$, the
size-weighted full-update ratio lies in $[0,1]$. If all indicators are zero, as in
Blockwise Muon, then $\rho_T^{\mathrm{w}}=0$.
If all indicators are one, as in Full Muon, then $\rho_T^{\mathrm{w}}=1$.
The monitoring collectives synchronize scalar statistics and are not
included in the full-orthogonalization communication.
\end{proof}

\end{document}